\documentclass{article}
\usepackage{ijcai22}

\usepackage{times}
\usepackage{makecell}
\usepackage{soul}
\usepackage{enumitem}
\usepackage{url}
\usepackage[hidelinks]{hyperref}
\usepackage[utf8]{inputenc}
\usepackage[small]{caption}
\usepackage{graphicx}
\usepackage{subfigure}
\usepackage{amsmath}
\usepackage{amsthm}
\usepackage{booktabs}
\usepackage{algorithm}
\usepackage{algorithmic}
\usepackage{listings}
\usepackage{color}
\usepackage{multirow}
\usepackage{bbding}

\usepackage{amssymb}
\usepackage{booktabs}

\newlength\myindent
\definecolor{dkgreen}{rgb}{0,0.6,0}
\definecolor{gray}{rgb}{0.5,0.5,0.5}
\definecolor{mauve}{rgb}{0.58,0,0.82}

\title{Augmenting Anchors by the Detector Itself}

\author{
}

\author{
Xiaopei Wan$^1$\footnotemark[1]\and
Guoqiu Li$^2$\footnotemark[1]\and
Yujiu Yang$^2$\And
Zhenhua Guo$^3$\footnotemark[2]\\
\affiliations
$^1$Ant Group\\
$^2$Tsinghua University\\
$^3$Alibaba Group\\
\emails
ranghou.wxp@alibaba-inc.com,
lgq20@mails.tsinghua.edu.cn,
yang.yujiu@sz.tsinghua.edu.cn,
cszguo@gmail.com
}

\begin{document}

\maketitle

\renewcommand{\thefootnote}{\fnsymbol{footnote}}
\footnotetext[1]{These authors have contributed equally to this work.}
\footnotetext[2]{Corresponding authors.}

\begin{abstract}
  Usually, it is difficult to determine the scale and aspect ratio of anchors for anchor-based object detection methods. Current state-of-the-art object detectors either determine anchor parameters according to objects' shape and scale in a dataset, or avoid this problem by utilizing anchor-free methods, however, the former scheme is dataset-specific and the latter methods could not get better performance than the former ones. In this paper, we propose a novel anchor augmentation method named AADI, which means {\bf A}ugmenting {\bf A}nchors by the {\bf D}etector {\bf I}tself. AADI is not an anchor-free method, instead, it can convert the scale and aspect ratio of anchors from a continuous space to a discrete space, which greatly alleviates the problem of anchors' designation. Furthermore, AADI is a learning-based anchor augmentation method, but it does not add any parameters or hyper-parameters, which is beneficial for research and downstream tasks. Extensive experiments on COCO dataset demonstrate the effectiveness of AADI, specifically, AADI achieves significant performance boosts on many state-of-the-art object detectors (eg. at least +2.4 box AP on Faster R-CNN, +2.2 box AP on Mask R-CNN, and +0.9 box AP on Cascade Mask R-CNN). We hope that this simple and cost-efficient method can be widely used in object detection. Code and models are available at \hyperlink{https://github.com/WanXiaopei/aadi}{https://github.com/WanXiaopei/aadi}.
\end{abstract}
\renewcommand{\thefootnote}{\arabic{footnote}}
\section{Introduction}

Object detection, which needs to localize and recognize each object in images or videos, is a fundamental and critical problem in computer vision. 
With the development of deep learning, current object detectors are usually based on convolution neural 
networks, like FCOS~\cite{tian2019fcos}, RepPoints~\cite{yang2019reppoints,chen2020reppointsv2}, TOOD~\cite{feng2021tood},  Faster R-CNN\cite{ren2016faster}, DETR~\cite{carion2020end}, DeFCN~\cite{wang2021end}, and Sparse R-CNN~\cite{sun2020sparse}. They can be roughly divided into anchor-free or anchor-based methods
based on their ways to generate detection results. Anchor-based methods firstly place a lot of anchors with different scales and aspect ratios,
and then refine these anchors to generate proposals or detection results, while anchor-free methods directly generate them from feature maps.

\begin{figure}[t]
  \centering
  \subfigure[]{\label{aadi_for_rpn:a}\includegraphics[width=0.31\linewidth]{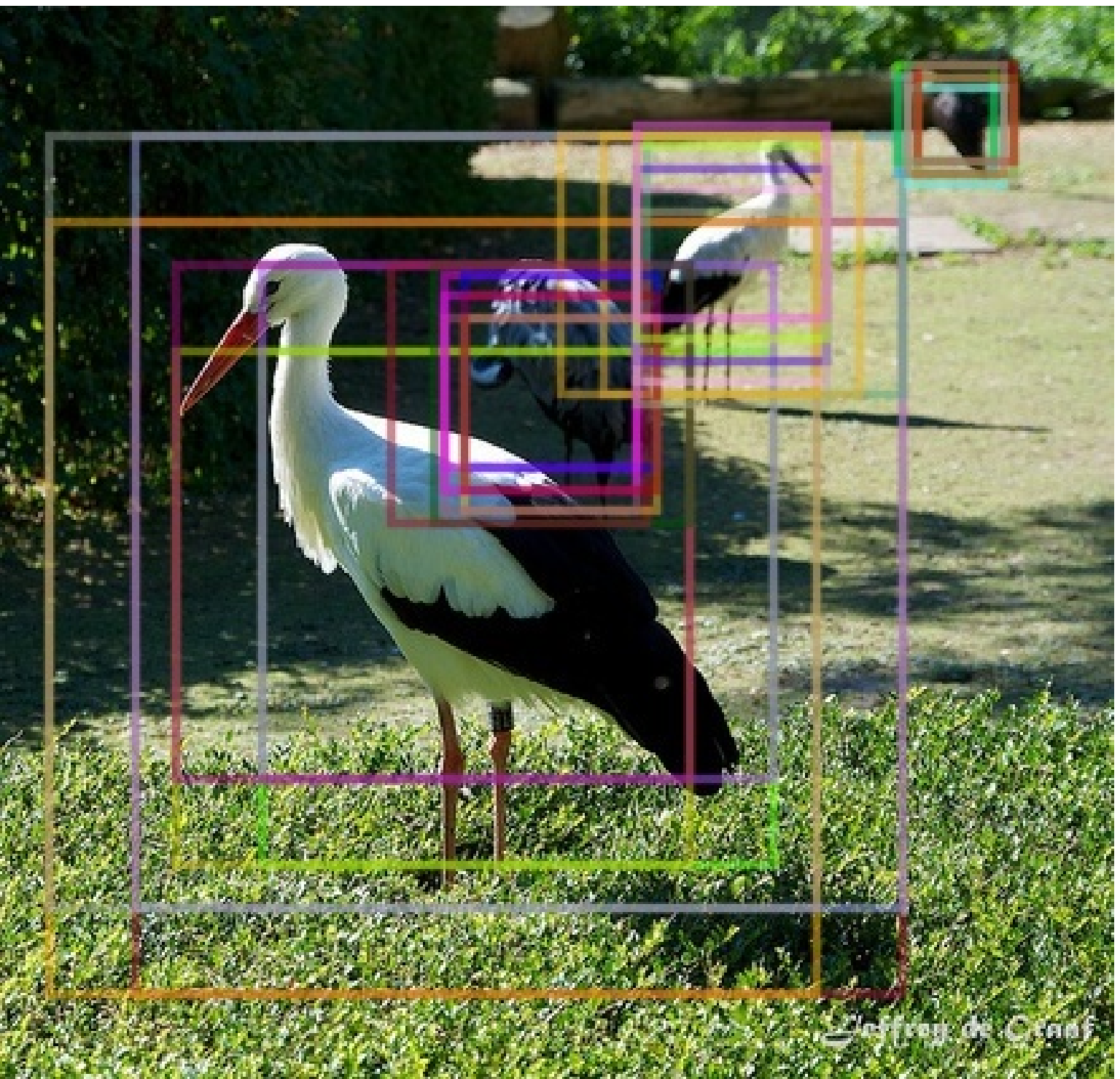}}
  \subfigure[]{\label{aadi_for_rpn:b}\includegraphics[width=0.31\linewidth]{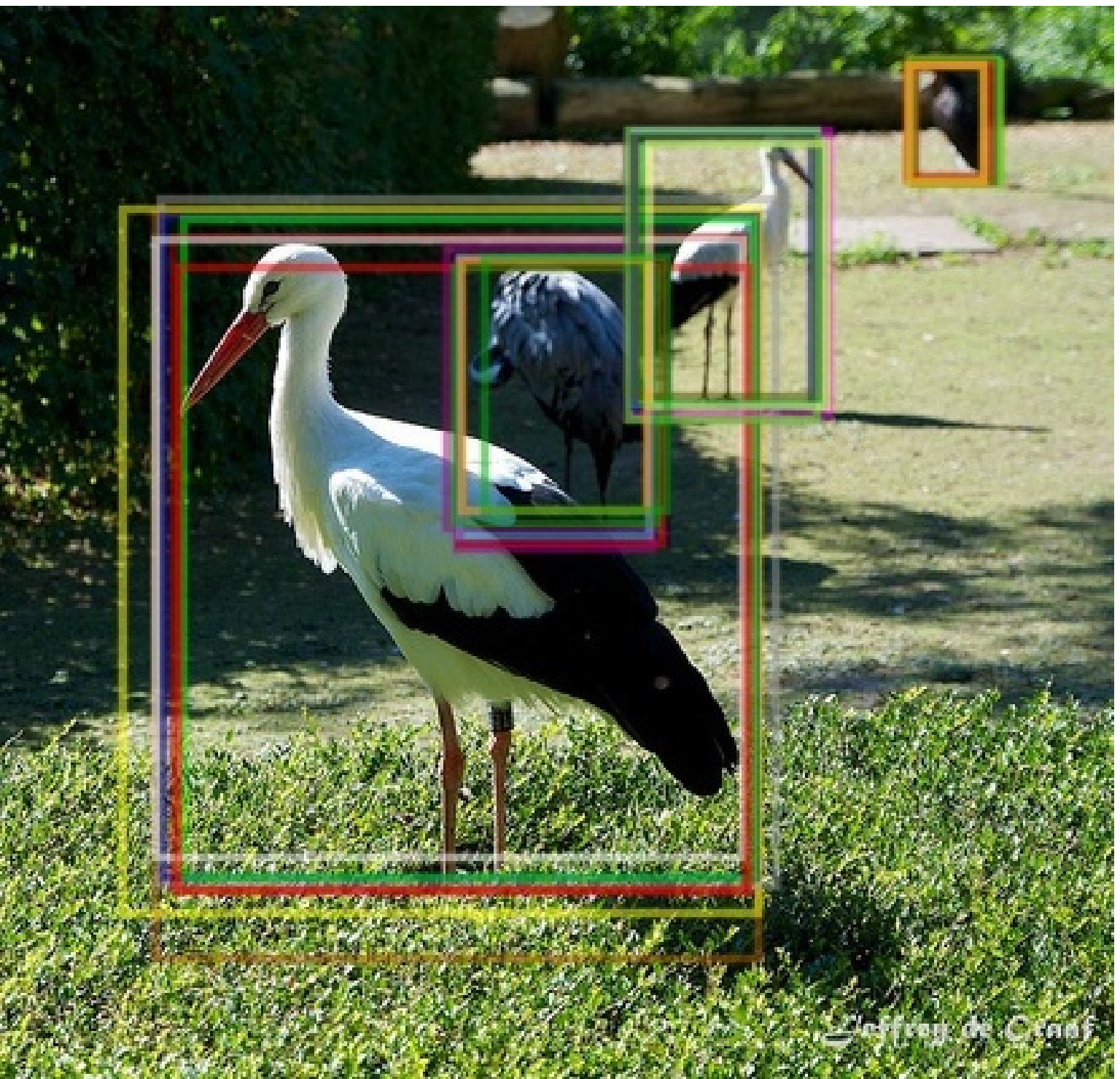}}
  \subfigure[]{\label{aadi_for_rpn:c}\includegraphics[width=0.31\linewidth]{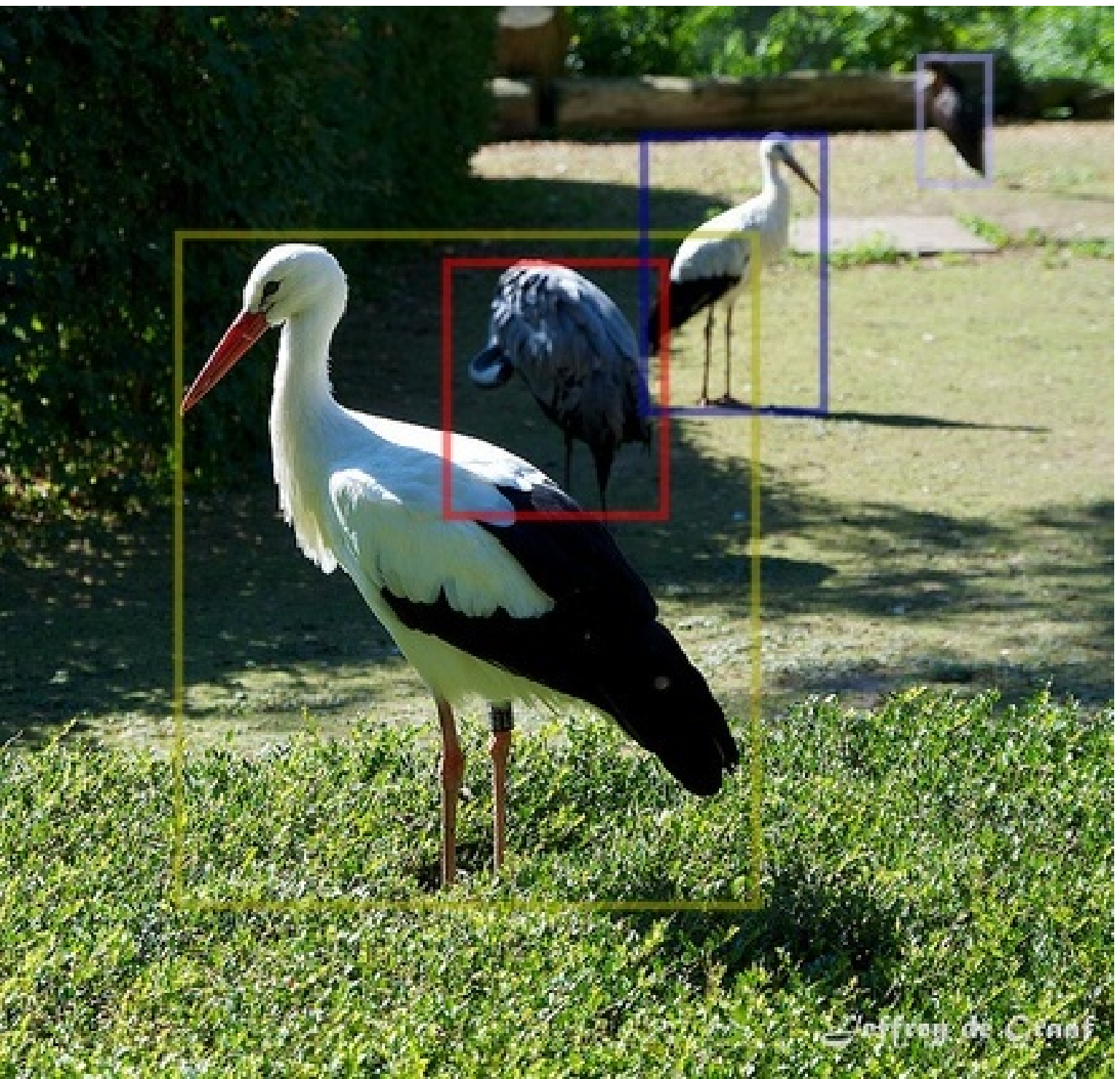}}
  \caption{Visualization of AADI for RPN. (a) The original anchors of standard RPN. (b) The anchors augmented by AADI. (c) The proposals generated by AADI-RPN}
  \label{aadi_for_rpn}
\end{figure}

The anchor-free methods were once widely regarded as the standard paradigm for future object detectors. However, ATSS~\cite{zhang2020bridging} has demonstrated that anchor-free methods are not more effective than anchor-based ones. For example, with same tricks and appropriate strategies for judging positive and negative samples of hand-designed anchors, the  anchor-based ATSS can perform better than anchor-free FCOS. Based on ATSS, PAA~\cite{paa-eccv2020} proposes a method of judging positive and negative samples by Gaussian Mixture Model and achieves large improvements, which further proves the conclusion mentioned above. Both of these two models use a method to estimate the critical point of the distribution of positive and negative anchors. However, even if the best critical point is estimated, there is no guarantee that the positive and negative samples will be completely separated, and placing more hand-designed anchors can not solve this problem. Therefore, how to improve the quality of anchors has become an important problem.\footnotemark[1]
\footnotetext[1]{More details can be seen from appendix.} 

\begin{figure}[t]
  \centering
  \subfigure[$d$=1.]{\label{anchor_dilation:a}\includegraphics[width=0.3\linewidth]{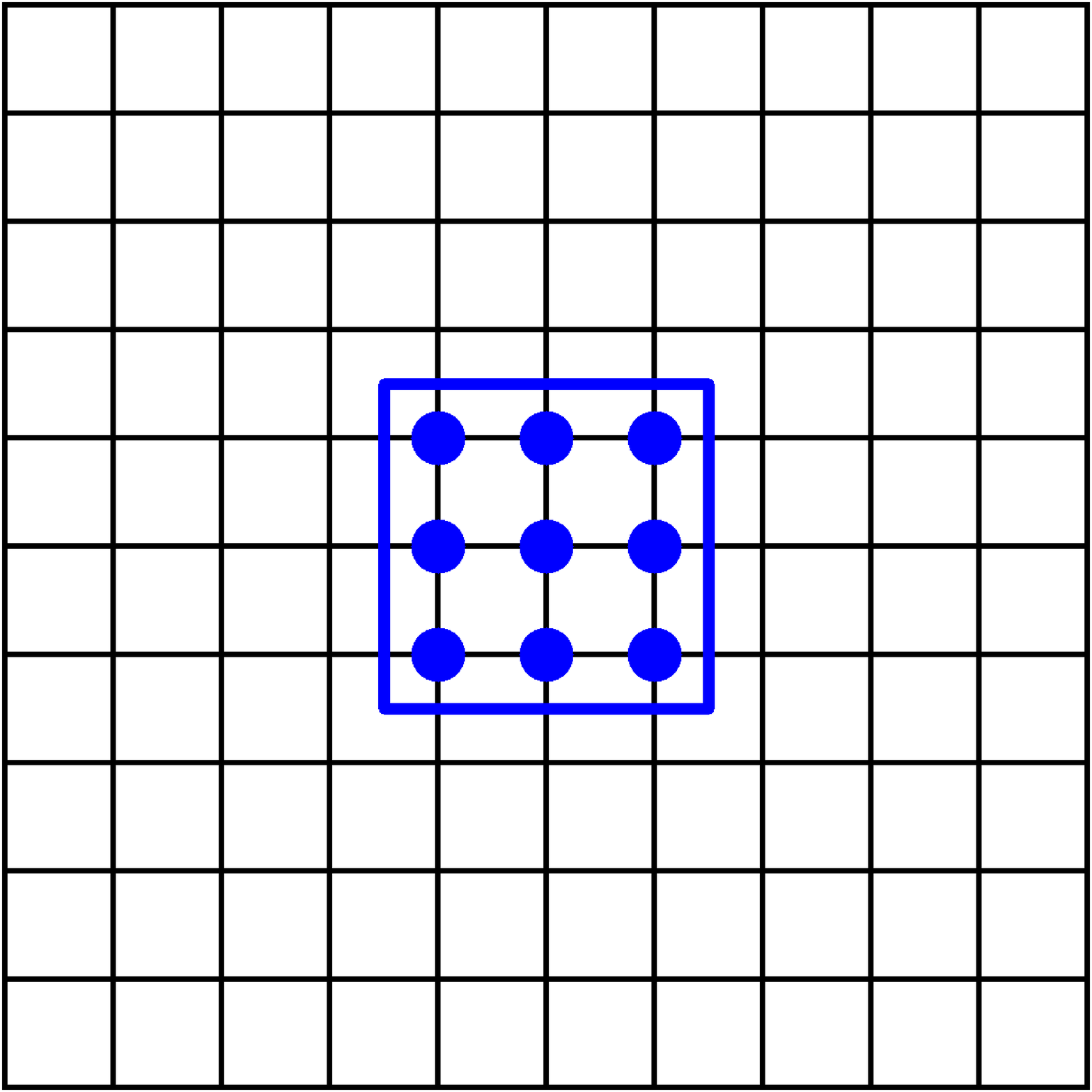}}
  \hspace{0.03\linewidth}
  \subfigure[$d$=2.]{\label{anchor_dilation:b}\includegraphics[width=0.3\linewidth]{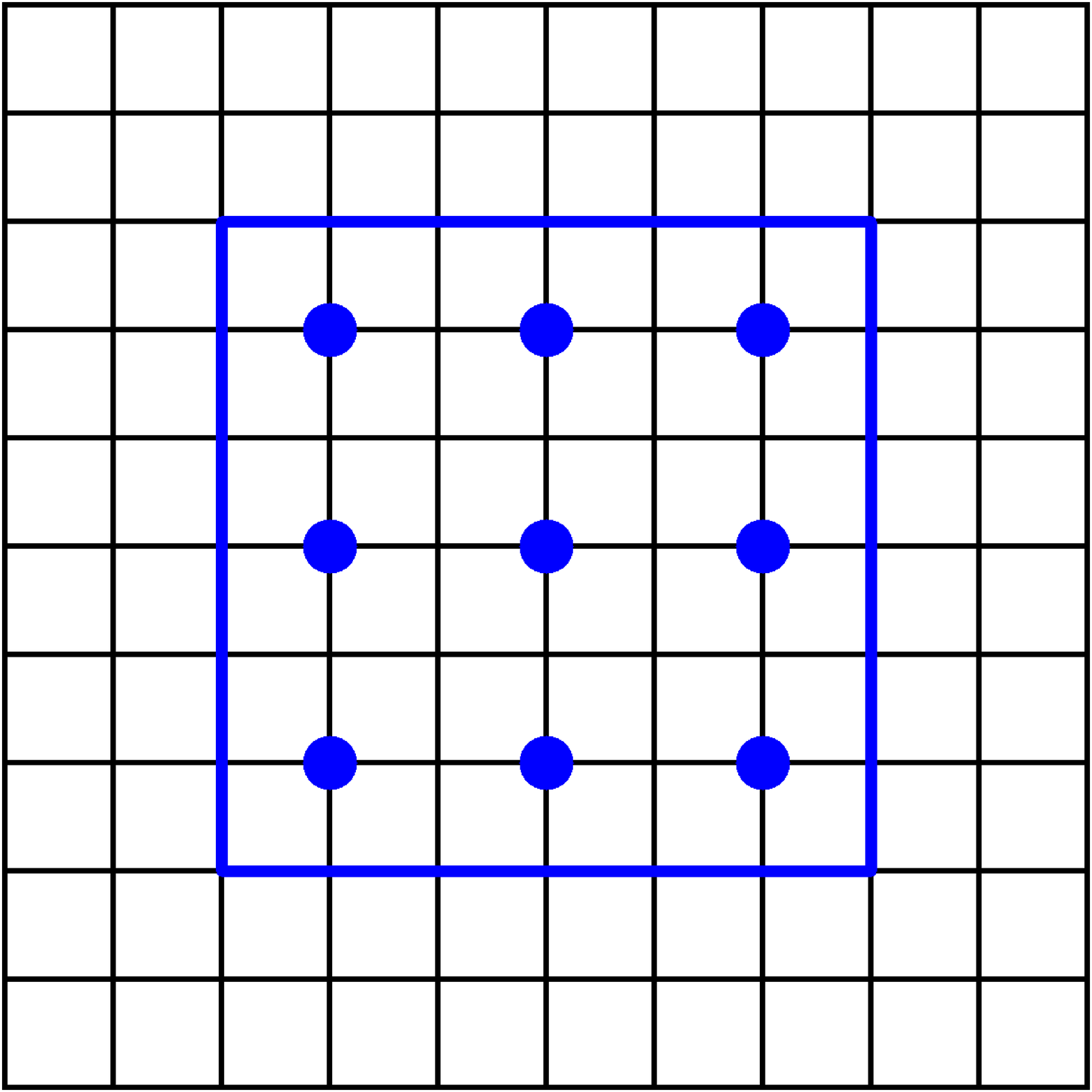}}
  \hspace{0.03\linewidth}
  \subfigure[$d$=3.]{\label{anchor_dilation:c}\includegraphics[width=0.3\linewidth]{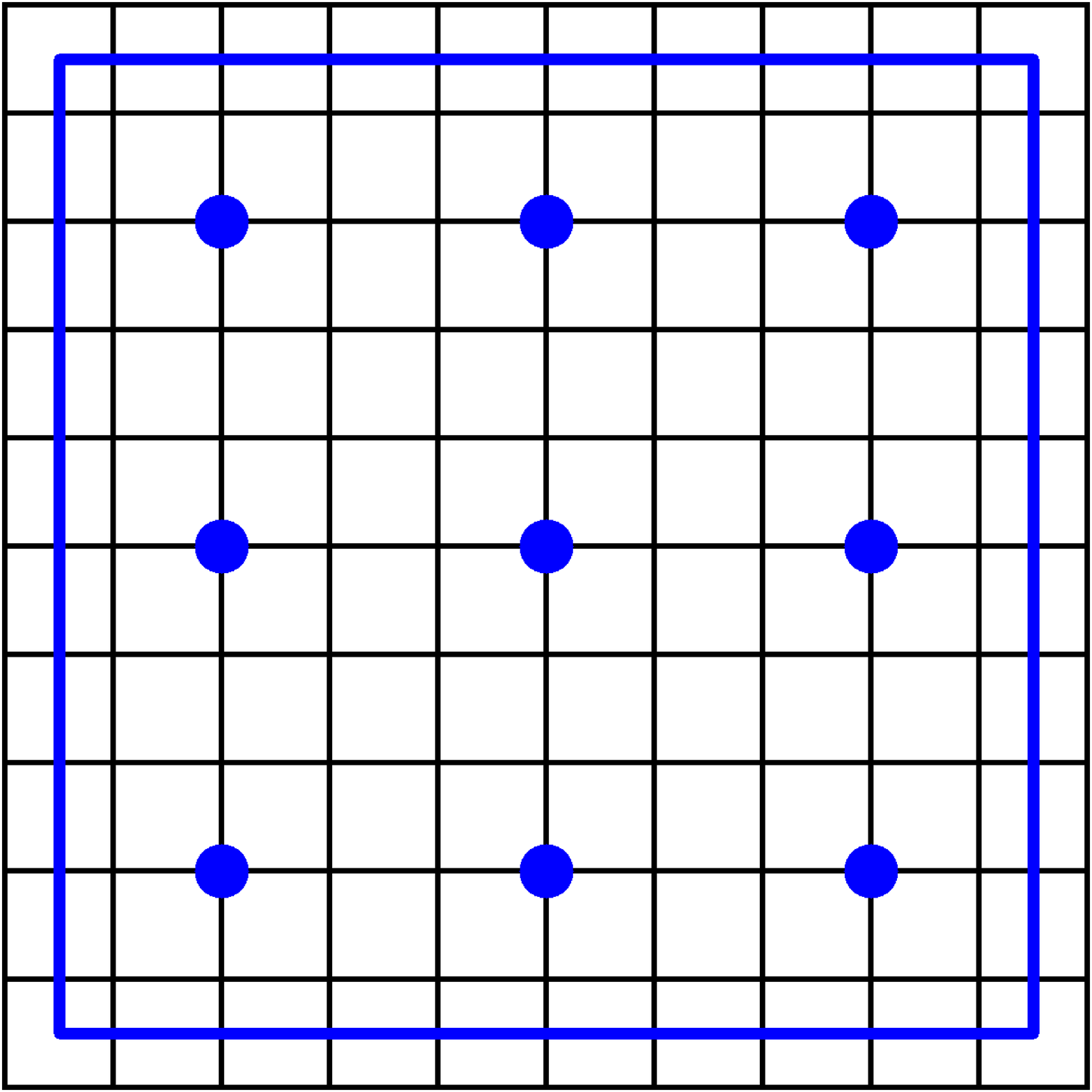}}
  \caption{Anchors under different dilations, and the kernel of the convolution layer is fixed to 3*3.}
  \label{anchor_dilation}
\end{figure}

In current anchor-based object detectors' paradigm, high-quality anchors usually lead to better performance, because they make it easier to separate positive and negative anchors, which is important to improve the recall rate of ground truth boxes and reduce the number of false positive predictions. 
In standard RPN, anchors with an IoU greater than 0.7 are marked as positive anchors, those with an IoU below 0.3 are marked as negative anchors, and others are not considered in training. Thus, placing more anchors with different aspect ratios and scales will improve the performance of RPN. However, since the aspect ratio and scale can be any positive number, it makes the anchor designation to be a non-trivial problem. Thus, some methods try to learn high-quality anchors by a neural network module, like AttratioNet~\cite{gidaris2016attend}, GA-RPN, and Cascade RPN.
All these methods can be viewed as gradient-based anchor augmentation methods, and their overall training and inference overheads almost double due to their strategy of dense prediction and a large amount of memory access in feature alignment.

In this paper, we propose a gradient-free anchor augmentation method named AADI, which means {\bf A}ugmenting {\bf A}nchors by the {\bf D}etector {\bf I}tself. AADI can be applied to both two-stage and single-stage methods, and Figure~\ref{aadi_for_rpn} shows an example of AADI for RPN~\cite{ren2016faster}. AADI comprises two processes, including augmentation and refinement process. During training process, AADI first uses RPN to augment hand-designed anchors to improve their quality in augmentation process, and then uses the augmented anchors and ground truth to train the parameters of RPN in refinement process. Intuitively, AADI can be viewed as a variational EM algorithm which is typically used to solve the problems with latent variables, and the augmentation and refinement process are corresponding to its E-step and M-step, respectively. Here, we iterate it only once to improve the computational efficiency. Moreover, in the augmentation process, AADI uses dilated convolution~\cite{yu2015multi} to augment each hand-designed anchor, and in the refinement process, it employs RoI Align~\cite{he2017mask} to extract features for augmented anchors and uses an equivalent fully connected layer to refine these anchors. Note that the dilated convolution and fully connected operations share parameters. 

In order to make RPN's parameters be reused in these two processes, we need to carefully design the dilation and kernel size of convolution layer to ensure that the convolution layer and fully connected layer could obtain same input feature for a specific anchor. As  Figure~\ref{anchor_dilation} shows, supposing that the dilation of a convolution layer is $d$, then the process of a 3$\times$3 convolution layer is same as that of an RoI Align for a (3$\times d$)$\times$(3$\times d$) anchor. Furthermore, the operation of a 3$\times$3 convolution layer for each anchor is equal to a fully connected layer whose output dimension is 9. Thus, unlike the ineffective masked deformable convolution layer in GA-RPN or adaptive convolution layer in Cascade RPN which is used for tackling the feature misalignment problem of hand-designed and learned anchors, AADI avoids this problem by determining the scale and aspect ratio of anchors according to the dilation and kernel size of convolution operation. 
Consequently, the hyper-parameters of AADI are kernel size and dilation of convolution layer, that is, AADI converts the scale and aspect ratio of anchors from a continuous space to a discrete space, which greatly alleviates the problem of anchor designation.

During training, the augmentation process does not need to calculate gradient, which makes the training speed of AADI fast. During inference, AADI only takes top 2000 augmented anchors for further refinement, which makes the inference speed fast. Extensive experiments on MS COCO dataset show that AADI can efficiently improve the detection performance of anchor-based object detectors, specifically, AADI achieves at least 2.4 box AP improvements on Faster R-CNN, 2.2 box AP improvements on Mask R-CNN, 1.8 box AP improvements on RetinaNet, and 0.9 box AP improvements on Cascade Mask R-CNN. And it runs only slightly slower than standard RPN. Our main contributions are as follows:

\begin{itemize}[itemsep=0pt,topsep=0pt,parsep=0pt,leftmargin=9.3pt]
 \item We propose a novel and effective method, AADI, which augments anchors by the detector itself and can significantly improve anchors' quality. It can be applied on all anchor-based object detectors without adding any parameters or hyper-parameters.
 \item AADI determines the scale and aspect ratio of anchors according to the kernel and dilation of a convolution layer, and it determines the aspect ratio according to the shape of the kernel, which greatly eases the anchor design difficulty.
 \item We carry out experiments on MS COCO dataset to verify the effectiveness of AADI. With only little extra computation costs, AADI achieves significant performance improvements on many representative object detectors with different backbones.
\end{itemize}


\section{Related Works}

Current object detection methods can be roughly categorized into two classes, one is anchor-based method and the other is anchor-free method. Anchor-based methods firstly place a lot of hand-designed anchors on the feature maps, and then perform classification and box regression on these anchors. Therefore, better anchors can greatly alleviate the difficulty of these two tasks. AttratioNet~\cite{gidaris2016attend} uses ARN to iteratively refine the hand-designed anchors several times. However, ARN needs to employ RoI Pooling to extract features for each anchor, and it is not a lightweight network, which makes its training and inference speed slow. GA-RPN~\cite{wang2019region} first predicts a suitable anchor for each position, and then uses the predicted anchors to train RPN. Cascade RPN~\cite{cai2018cascade} stacks two RPNs together to improve the quality of proposals. Its first RPN is used to refine the hand-designed anchors without classification, and the second one plays the same role as a standard RPN. GA-RPN and Cascade RPN have effectively improved the quality of proposals by improving the quality of anchors. However, they both need to use DCN\cite{dai2017deformable} or its variants to align the features and anchors, and there is a lot of memory access during this operation, resulting in slower training and inference speed. In this paper, we first filter a small set of augmented anchors, and then use RoI Align to extract the features for them, which can significantly reduce the memory access.

Compared with anchor-free methods, anchor-based methods are more intuitive, and they do not need to determine which ground truth should be placed on which level of pyramid features. However, FCOS~\cite{tian2019fcos} finds that determining the scale and aspect ratio of anchors is not a trivial issue, and these two parameters have a great influence on the performance. Therefore, FCOS proposes an anchor-free paradigm to generate the final results. For each position which is located in the center region of the ground truth, it directly predicts the distance from the position to the four corners of the ground truth. RepPoints~\cite{yang2019reppoints} is also an anchor-free method. It no longer directly predicts bounding boxes, but predicts a set of representative points for each ground truth, and then groups them to bounding boxes. DETR and Deformable DETR~\cite{zhu2020deformable} use Transformer to directly predict the detection results to avoid the anchor designation problem.

\renewcommand{\algorithmicrequire}{\textbf{Input:}} 
\renewcommand{\algorithmicensure}{\textbf{Output:}}
\begin{algorithm}[t]
\caption{Pseudocode of AADI}
\label{alg_aadi_formula}
\begin{algorithmic}[1]
\REQUIRE ~~\\
    $\mathcal{A}$ is a set of hand-designed anchors;\\
    $\mathcal{A}_a$ is a set of augmented anchors;\\
    $\mathcal{G}$ is a set of ground truth boxes on the image;\\
    $\mathcal{F}$ is feature maps obtained by backbone;
    $f$ is RPN;\\
    $\theta$ is a set of paramters of RPN;\\
    $\epsilon$ is learning rate;\\
\ENSURE ~~ \\
    $\mathcal{P}$ is a set of proposals;\\
\WHILE{not converge}
    \STATE $\mathcal{A}_a \gets f(\mathcal{A}, \mathcal{F}, \theta)$
    \STATE $\mathcal{P} \gets f(\mathcal{A}_a, \mathcal{F}, \theta)$
    \STATE $\mathcal{L} \gets loss(\mathcal{P}, \mathcal{G})$
    \STATE $\bigtriangleup_{\theta} \gets \frac{\partial{\mathcal{L}}}{\partial{f(\mathcal{A}_a, \mathcal{F}, \theta)}} \times \frac{f(\mathcal{A}_a, \mathcal{F}, \theta)}{\partial{\theta}}$
    \STATE $\theta \gets \theta - \epsilon \times \bigtriangleup_{\theta}$
\ENDWHILE
\end{algorithmic}
\end{algorithm}

Although the hyper-parameter adjustment of anchor-free methods is simpler, ATSS and PAA demonstrate that anchor-based methods can significantly outperform anchor-free methods with suitable anchor assignment strategies. However, the anchors of both ATSS and PAA are predefined, although both ATSS and PAA are very robust to the scale and aspect ratio of anchors, they are hard to further improve the performance by placing more anchors with different scales and aspect ratios. However, AADI can adaptively augment the hand-designed anchors to improve their quality.

\begin{figure*}[t]
  \centering
  \subfigure[RPN.]{
  \begin{minipage}[t]{0.40\linewidth}
  \centering
  \label{rpn:a}
  \includegraphics[width=0.9\linewidth]{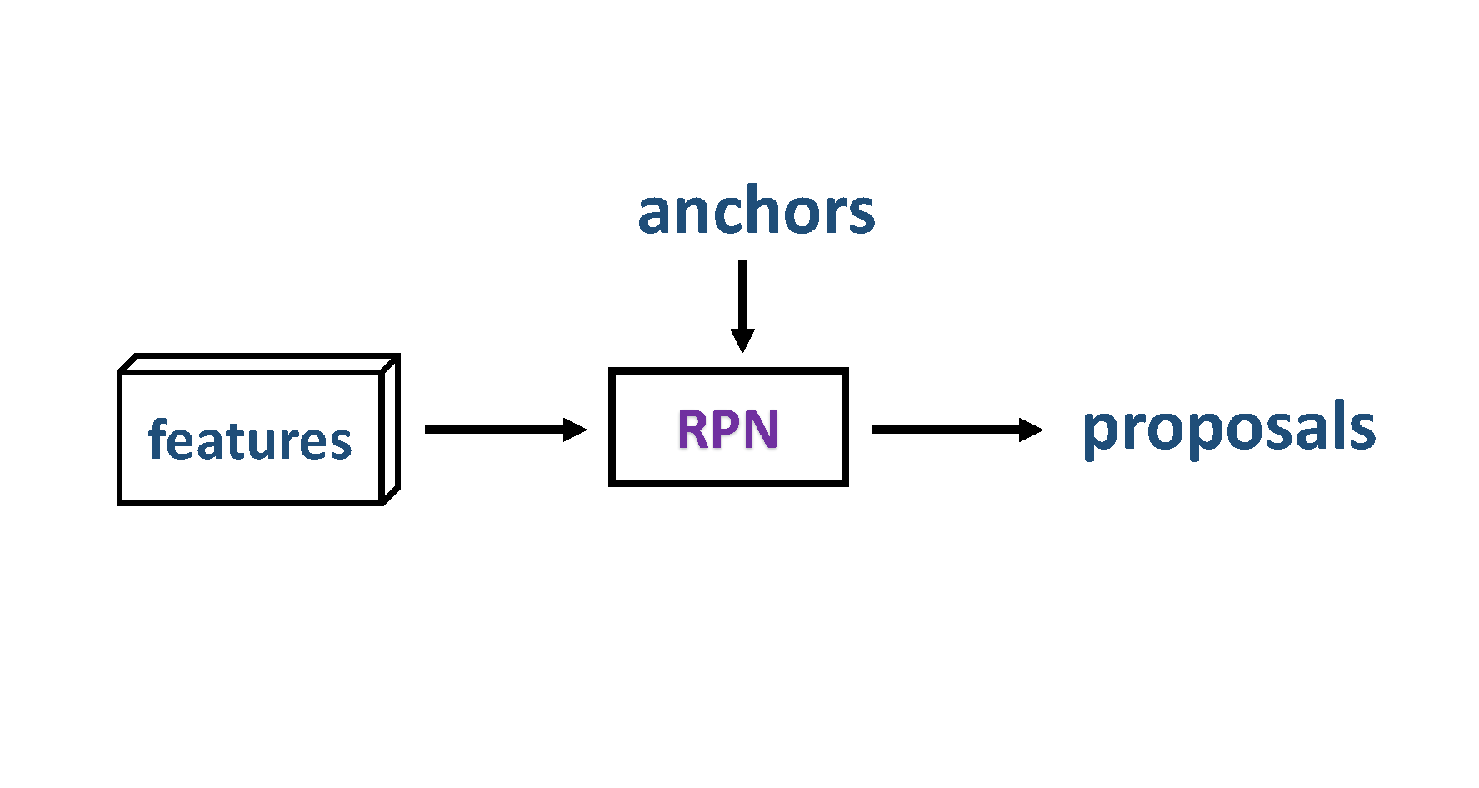}
  \end{minipage}
  }
  \subfigure[AADI-RPN.]{
  \begin{minipage}[t]{0.40\linewidth}
  \centering
  \label{aadi-rpn:b}
  \includegraphics[width=0.9\linewidth]{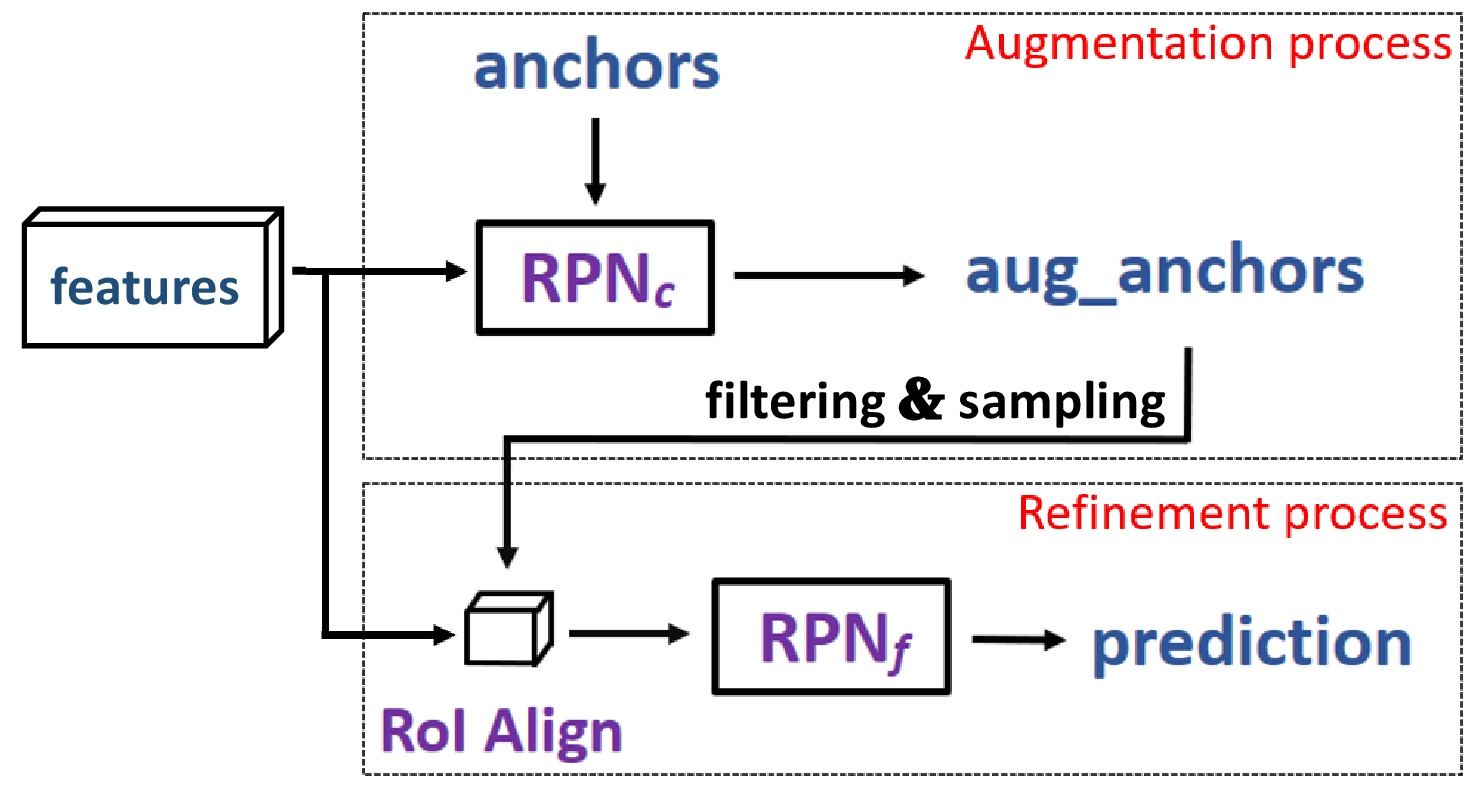}
  \end{minipage}
  }

  \subfigure[Box subnet of RetinaNet.]{
  \begin{minipage}[t]{0.45\linewidth}
  \centering
  \label{retinanet:c}
  \includegraphics[width=0.9\linewidth]{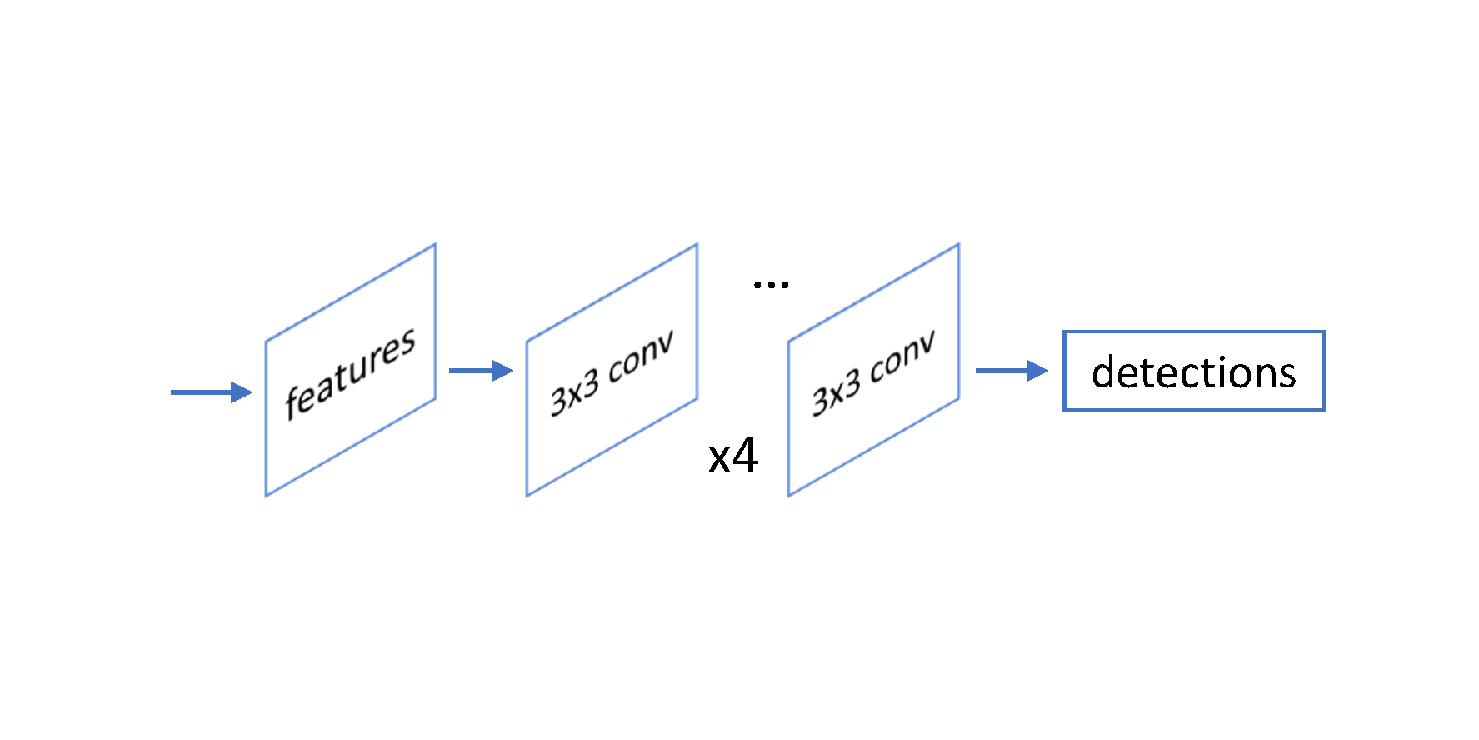}
  \end{minipage}
  }
  \subfigure[Box subnet of AADI-RetinaNet.]{
  \begin{minipage}[t]{0.50\linewidth}
  \centering
  \label{aadi-retinanet:d}
  \includegraphics[width=0.9\linewidth]{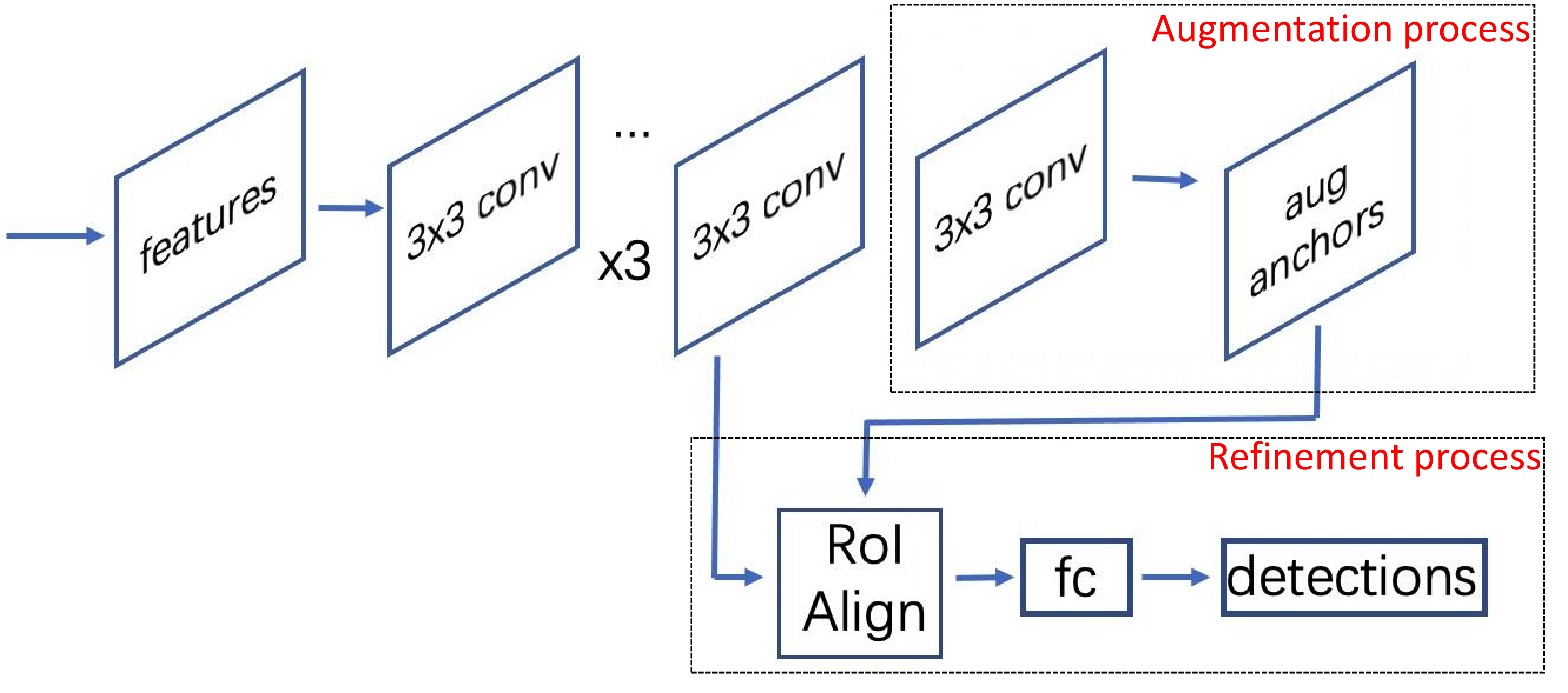}
  \end{minipage}
  }
  
  \caption{The pipeline of AADI for RPN and RetinaNet.}
  \label{AADI}
\end{figure*}

\section{Method}

\subsection{AADI}
\label{overview}


High-quality anchors are critical for all anchor-based object detectors, so, how to adaptively generate better anchors is an important factor. Unlike previous gradient-based methods, AADI just uses the detector itself to augment the hand-designed anchors without gradient, and it can be used in both two-stage and single-stage object detectors. The overall pipeline of AADI for a common two-stage object detector is shown in Figure~\ref{aadi-rpn:b}, and its Pseudocode is shown in Algorithm~\ref{alg_aadi_formula}. Compared with AttratioNet, AADI does not need to calculate gradient during anchor augmentation. And compared with the IBBR({\bf I}terative {\bf B}ounding {\bf B}ox {\bf R}egression) of IoU-Net~\cite{jiang2018acquisition}, which has a similar inference pipeline. The key difference is that AADI works on both training and inference period, while IBBR only works on inference. Taking AADI for two-stage object detectors as an example, AADI comprises two processes, including an augmentation process and a refinement process, and the details are illustrated as follows.

\noindent\textbf{Augmentation process.}

As shown in Figure~\ref{aadi-rpn:b}, AADI first gets the input of anchors and feature maps from the backbone, and then uses RPN$_c$ to augment the anchors. RPN$_c$ is same as the standard RPN, and it does not need to calculate the gradient during training. As most of augmented anchors are either redundant samples or negative samples, so we remove them by Non-Maximum Suppression (NMS) and keep only 2000 anchors with the highest score to improve the model's efficiency. As shown in Figure~\ref{aadi_for_rpn}, the augmented anchors have a higher Intersection over Union (IoU) than the original anchors.

Note that there can be multiple RPN$_c$ with different dilations, and these RPN$_c$s can share parameters.

\noindent\textbf{Refinement process.}

The refinement process gets input of augmented anchors from the augmentation process and feature maps from the backbone, and then uses RPN$_f$ to predict proposals. RPN$_f$ and RPN$_c$ have the same parameters, but they are implemented in different ways. RPN$_c$ is same as the standard RPN, and it uses convolution to get augmented anchors, while RPN$_f$ is implemented in a fully connected way. In addition, RPN$_f$ employs RoI Align to extract features for each augmented anchor. Note that different RPN$_c$s are to correspond to different RPN$_f$s if their parameters are not sharing. Moreover, RPN$_f$ needs to calculate gradient during training, so RPN$_c$ will also be trained indirectly.

The key problem of ensuring that this paradigm can work is the features obtained by the two RPNs should be same. For RPN$_c$, the features for each anchor are determined by the kernel size and dilation of a convolution layer, while for RPN$_f$, the features are determined by RoI Align. So, there are two conditions to meet. First, the output size of RoI Align needs to be same as the kernel size of the convolution layer, which is easy to meet. Second, the dilation and the scale of each anchor must be carefully designed to align the RoI, which is hard to meet. 

To address this issue, we propose to set scale and aspect ratio of hand-designed anchors by  Eq.~\ref{eq:scale_determiniation} and Eq.~\ref{eq:aspect_ratio_determiniation}, respectively. Where $m$ and $n$ is height and width of kernel size of a convolution layer, and $d$ is the dilation. In Figure~\ref{anchor_dilation}, we visualize the anchors and their features sampled by a 3x3 convolution layer under different dilations, and it is easy to find that the features sampled by a 3$\times$3 convolution layer are same as the features extracted by a 3$\times$3 RoI Align. In this way, the aspect ratio of the anchors is also determined by the shape of the kernel. For example, for a 3$\times$3 convolution layer, if its dilation is set to 2, then the scale of the anchors is $3 \times 2=6$, and the aspect ratio is $\frac{3}{3}=1$. Since $m$, $n$ and $d$ all are must be integers, so the designation space of anchors is converted from a continuous space to a discrete space, which greatly alleviates the problem of anchor designation.

\begin{equation}
  scale = d \times \sqrt{m \times n}
  \label{eq:scale_determiniation}
\end{equation}

\begin{equation}
  aspect\_ratio = \frac{n}{m}
  \label{eq:aspect_ratio_determiniation}
\end{equation}

\subsection{Training}

If AADI is only trained on the augmented anchors, AADI's robustness to hand-designed anchors is relatively poor, which will reduce the performance of anchor augmentation. Therefore, during training, we first select an anchor with the largest IoU for each ground truth, then append them to the augmented anchors, which can guide the AADI to augment the hand-designed anchors, we call this strategy {\bf Anchor Guided}.

The number of positive augmented anchors is very large, if all of them are used directly, the memory will be exhausted. In addition, we found that the overlap of some augmented 
anchors is large. Therefore, we employ NMS to filter out the redundant anchors according to their IoUs with the ground truth, and the training strategy of the refinement process is same as standard RPN. Similar to RPN, AADI can be trained in an end-to-end manner using multi-task loss as follows:

\begin{equation}
  \mathcal{L}_{rpn} = \lambda \mathcal{L}_{reg} +  \mathcal{L}_{cls}
  \label{eq:multi_task}
\end{equation}

Where $\mathcal{L}_{reg}$ and $\mathcal{L}_{cls}$ are the box regression loss and the classification loss, respectively. And these two loss terms are balanced by $\lambda$. In the implementation, binary cross entropy loss and Smooth L1 Loss~\cite{girshick2015fast} are used as classification loss and box regression loss, respectively, and $\lambda$ is empirically set to 5.

\subsection{Inference}

During inference, the augmentation process is same as that of obtaining proposals by standard RPN, and the refinement process is just a two-layer fully connected layer which is used to refine the augmented anchors. Here, AADI only takes the top 2000 augmented anchors for refinement, while Cascade RPN needs to augment all regressed anchors, which leads to higher computation overheads.

\subsection{AADI-RetinaNet}

AADI can be naturally deployed in anchor-based single-stage object detectors. We only apply AADI on RetinaNet (AADI-RetinaNet) due to the limited time and resources. Since RetinaNet uses 4 multi-layer convolution networks to extract features for classification and regression tasks respectively, and the kernel of all convolution layers is 3x3, so we can use the last layer of the box subnet to extract features for the augmented anchors. The pipeline of AADI-ReitinaNet are shown in Figure~\ref{aadi-retinanet:d}, and the training process of AADI-RetinaNet is similar to AADI-RPN, However, during inference, since NMS is relatively time-consuming,  we only select the top 2000 augmented anchors without NMS to do further refinement. Here, the IoU threshold of NMS of post-processing is set to 0.6. Furthermore, Group Normalization~\cite{wu2018group} is used to make the training process smoother.

\section{Experiments}

\subsection{Dataset and Evaluate Metrics}

Our experiments are implemented on the challenging Microsoft COCO 2017~\cite{lin2014microsoft} dataset. It consists of 118k images for training, {\it train-2017} and 5k images for validation, {\it val-2017}. 
There are also 20k images without annotations for test in {\it test-dev}. 

We train all our models on {\it train-2017}, and report ablation studies on {\it val-2017} and final results on {\it test-dev}. In this paper, Average Recall(AR) is used to measure the quality of proposals, which is the average of recalls at different IoU thresholds, and AR$_x$ means recall at IoU threshold of $x$\%, AR$_s$, AR$_m$, and AR$_l$ denote the AR for small, medium, and large objects, respectively. The detection results are subjected to standard COCO-style Average Precision (AP) metrics which include AP (mean of AP over all IoU thresholds), AP$_{50}$, AP$_{75}$, AP$_s$, AP$_m$ and AP$_l$, where AP$_{50}$ and AP$_{75}$ are APs for IoU threshold of 50\% and 75\%, respectively, and AP$_s$, AP$_m$ and AP$_l$ are APs for small, medium, and large objects, respectively. Moreover, AP$^\text{box}$ and AP$^\text{mask}$ are denote AP for object detection and instance segmentation, respectively.

\subsection{Implementation Details}

All experiments are based on detectron2~\cite{wu2019detectron2}. And our codes are deployed on a machine with 8 Tesla V100-SXM2-16GB GPUs and Intel Xeon Platinum 8163 CPU. Our software environment mainly includes Ubuntu 18.04 LTS OS, CUDA 10.1, and PyTorch~\cite{paszke2017automatic} 1.6.0. 

Unless specially noted, the hyper-parameters of our model are following detectron2, and all models are based on FPN~\cite{lin2017feature}. For AADI-RPN, the dilations of RPN$_c$ is set to 2 and 4, and these two RPN$_c$ are not sharing parameters, moreover. For AADI-RetinaNet, the dilation is set to 2. Moreover, the strategy of Anchor Guided is used in all models.

As a common practice~\cite{girshick2014rich}, all network backbones are pretrained on ImageNet1k classification set~\cite{5206848}, and their pre-trained models are publicly available. With 1x~\cite{wu2019detectron2} training schedule and ResNet-50 (R-50) as backbone, it takes about 4.2 hours for our model to converge, and about 2.9 hours with Mixed Precision training~\cite{micikevicius2017mixed}. 

\subsection{Results}

\begin{table}
  \begin{center}
   \resizebox{0.45\textwidth}{18mm}{
      \begin{tabular}{lcccccccc}
          \toprule
          & & & & & & & \multicolumn{2}{c}{Speed}\\
          \cmidrule(r){8-9}
          Method & Backone & AR$_{100}$& AR$_s$ & AR$_m$ & AR$_l$ & AP$^\text{box}$ & Train(ms) & Test(FPS)\\
          \midrule
          AttratioNet & VGG-16    & 53.3 & 31.5 & 62.2 & 77.7 & 34.3 & - & -\\
          ZIP         & Inception & 53.9 & 31.9 & 63.0 & 78.5 & - & - & -\\
          C-RPN       & R-50      & 54.0 & 35.6 & 62.7 & 78.5 & - & - & -\\
          \midrule
          RPN         & R-50 & 45.7 & 31.4 & 52.7 & 61.1 & 37.9 & 186.0 & 31.3\\
          PAA         & R-50 & 49.7 & 31.4 & 59.4 & 68.3 & 39.3 & 342.6 & 31.4\\
          GA-RPN      & R-50 & 59.1 & 40.7 & 68.2 & 78.4 & 39.6 & 448.1 & 14.7\\
          Cascade RPN & R-50 & 61.1 & 42.1 & 69.3 & 82.8 & 40.4 & 436.4 & 15.1\\
          \midrule
          AADI-RPN    & R-50 & 55.3 & 35.4 & 62.7 & 79.7 & 41.3 & 170.2 & 25.6\\
          \bottomrule
      \end{tabular}}
  \end{center}
  \caption{Region proposal results on COCO {\it val-2017}. All results are trained with 1x schedule.}
  \label{proposal_recall}
\end{table}

{\bf Region Proposal Performance.} We first apply AADI on RPN (AADI-RPN) of Faster R-CNN to verify its performance on region proposal, and the model is trained on COCO {\it train-2017} with 1x training schedule.
Since it is difficult to reproduce the performance of AttratioNet, ZIP\cite{li2019zoom} and C-RPN~\cite{fan2019siamese}, we only compare recall rate with published results. For other models, their performance is tested on the same machine, and the training time indicates the average time cost per batch, while the test time indicates the average inference speed on COCO {\it val-2017}. All region proposal results are shown in Table~\ref{proposal_recall}, and the results show that AADI-RPN, by a large margin, outperforms standard RPN in terms of AR under all settings, and it achieves the highest AP$^{\text{box}}$ over other models. Furthermore, although AADI-RPN has a lower recall rate than the current state-of-the-art Cascade RPN, its inference speed is about 70\% faster. For training time, AADI-RPN achieves the fastest training speed, even faster than RPN. This is mainly because only a small set of the augmented anchors are participated in the training process, and the way to compute the gradient is changed from convolution to fully connected operation, which significantly reduces the training overheads.

\begin{table*}
   \begin{center}
   \resizebox{0.78\textwidth}{25mm}{
      \begin{tabular}{ccc|cccccc|c}
          \hline
          
          Method & Backbone & AADI  & AP$^{\text{box}}$ & AP$_{50}^{\text{box}}$ & AP$_{75}^{\text{box}}$ & AP$^{\text{mask}}$ & AP$_{50}^{\text{mask}}$ & AP$_{75}^{\text{mask}}$   & Inference Speed(FPS)\\
          \hline
          \multirow{4}{*}{RetinaNet} & R-50 &  & 39.6 & 59.3 & 42.2 & - & - & -  & 22.2\\  
                 & R-50   & \checkmark & \textbf{41.4} & \textbf{59.3} & \textbf{45.2} & - & - & -  & 23.0\\\cline{4-10}
                 & Swin-T &            & 44.9          & \textbf{65.8} & 48.0          & - & - & -  & 18.6\\  
                 & Swin-T & \checkmark & \textbf{47.0} & 65.6          & \textbf{51.4} & - & - & & 18.1\\
          \hline
          \multirow{4}{*}{Faster R-CNN} & R-50 &  & 40.4 & 61.4 & 43.9 & - & - & - & 26.0\\  
                 & R-50   & \checkmark & \textbf{42.8}    & \textbf{61.4}    & \textbf{46.9}    & - & - & - & 18.7\\\cline{4-10}
                 & Swin-T &            & 45.0          & 67.2          & 49.1          & - & - & - & 19.6\\  
                 & Swin-T & \checkmark & \textbf{47.7} & \textbf{67.5} & \textbf{52.4} & - & - & - & 15.1\\
          \hline
          \multirow{4}{*}{Mask R-CNN}  & R-50 & & 41.0 & 61.5 & 44.9 & 37.2 & 58.6 & 39.9  & 22.8\\
            & R-50 & \checkmark   & \textbf{43.2} & \textbf{61.7} & \textbf{47.7} & \textbf{38.1} & \textbf{58.9} & \textbf{41.2}  & 15.8\\\cline{4-10}
            & Swin-T &            & 46.0          & \textbf{68.1} & 50.3          & 41.6          & \textbf{65.1} & 44.9  & 17.9\\
            & Swin-T & \checkmark & \textbf{48.4} & 67.6          & \textbf{53.2} & \textbf{42.8} & 65.0 & \textbf{46.6}  & 13.4\\
          \hline
         \multirow{4}{*}{\makecell{Cascade \\Mask R-CNN}} 
            & R-50 &  & 44.4 & 61.7 & 48.3 & 38.2 & 59.0 & 41.1  & 12.5\\
            & R-50   & \checkmark & \textbf{45.4}   &\textbf{62.4}  & \textbf{49.2} & \textbf{38.9} & \textbf{59.4} & \textbf{42.0} & 9.9 \\\cline{4-10}
            & Swin-T &            & 50.3          & 69.0 & 54.9 & 44.0 & 66.5 & 47.8  & 10.7\\
            & Swin-T & \checkmark & \textbf{51.2} & \textbf{69.2}  & \textbf{55.8} & \textbf{44.1} & \textbf{66.6} & \textbf{48.1} & 8.9\\
         \hline
      \end{tabular}}
    \end{center}
  \caption{Object detection results on COCO {\it val-2017}. All results are trained with 3x schedule.}
  \label{object_detection}
\end{table*}

{\bf Object Detection Performance.} We conducted extensive experiments on four typical object detection frameworks: RetinaNet, Faster R-CNN, Mask R-CNN, and Cascade Mask R-CNN, as well as two different backbone networks: R-50 and tiny Swin Transformer (Swin-T)~\cite{liu2021Swin} to verify the anchor augmentation capabilities of AADI. All detection results with 3x training schedule are shown in Table~\ref{object_detection}, and we can see that AADI achieves significant performance boosts in all detectors with different backbones, which demonstrates its strong robustness. In particular, when using Swin-T as the backbone, we observe +2.1 AP$^\text{box}$, +2.7 AP$^\text{box}$, +2.4 AP$^\text{box}$, and +0.9 AP$^\text{box}$ gains over RetinaNet, Faster R-CNN, Mask R-CNN, and Cascade Mask R-CNN, respectively. Furthermore, their AP$^\text{mask}$ are also improved. It is worth noting that the Swin Transformer based object detectors are currently state-of-the-art, while AADI still achieves significant performance boosts with only little extra computation costs, which reflects the efficiency of AADI. Meanwhile, we can find that the performance gains on AP$_{50}^{\text{box}}$ are small, while large gains are obtained on AP$_{75}^{\text{box}}$, similar finding could be found for GA-RPN and Cascade RPN. Therefore, the main advantage of AADI is to improve the quality of anchors, thereby improving the detection performance.

\section{Ablation Study}

We conduct our ablation study on COCO {\it train-2017}, and report the results on COCO {\it val-2017} dataset. All models' backbone is R-50 unless specially noted.


It can be seen from Eq.~\ref{eq:scale_determiniation} and Eq.~\ref{eq:aspect_ratio_determiniation} that the scale and aspect ratio of anchors is determined by the kernel and dilation of a convolution layer. For RPN, since the kernel is usually fixed to 3*3, so we only change the scale of anchors by adjusting the dilation. Here, we test the region proposal performance of RPN with dilation of 2, 3 and 4, respectively. Furthermore, the effects of the proposed training strategy of {\bf Anchor Guided} in augmentation process are also tested. All results are shown in Table~\ref{ablation_proposal_recall}.

From Table~\ref{ablation_proposal_recall}, we can conclude that larger dilation, worse performance of small objects, and better performance of large objects, and vice versa. This is because the scale of anchors will become larger as the dilation becomes larger, and large anchors are beneficial for large objects, but harmful for small objects during training. Furthermore, the strategy of Anchor Guided can consistently improve the performance of AADI-RPN with different dilations, especially for large objects. Compared with standard RPN, AADI-RPN consistently gets better performance by a large margin under different settings, which indicates that AADI is an effective method.

\begin{table}
  \begin{center}
  \resizebox{0.45\textwidth}{14mm}{
      \begin{tabular}{ccccccccc}
          \toprule
          Dilation & Anchor Guided & AR$_{100}$ & AR$_{1000}$ & AR$_s$ & AR$_m$ & AR$_l$\\
          \midrule
          2  &            & 54.8 & 64.7 & 39.0 & 63.1 & 70.6\\
          2  & \checkmark & 56.3 & 66.7 & 39.5 & 64.9 & 73.4\\
          3  &            & 53.7 & 64.0 & 35.4 & 62.1 & 73.9\\
          3  & \checkmark & 55.6 & 67.6 & 36.1 & 64.3 & 77.6\\
          4  &            & 52.2 & 60.5 & 30.9 & 61.3 & 76.6\\
          4  & \checkmark & 54.4 & 65.5 & 33.0 & 63.7 & 78.9\\
          \midrule
          \multicolumn{2}{c}{RPN} & 45.7 & 58.0 & 31.4 & 52.7 & 61.1\\
          \bottomrule
      \end{tabular}}
  \end{center}
  \caption{Region proposal results of AADI-RPN with different settings. All results are trained with 1x schedule.}
  \label{ablation_proposal_recall}
\end{table}

We find that there is not much difference in the performance of AADI-RPN with different dilations. Therefore, we test detection performance to determine which dilation or its combination is the optimal choice. We further test whether the parameters are shared or not when using different dilation combinations. All detection results are shown in Table~\ref{ablation_object_detection}, and the strategy of Anchor Guided is used for all tests. 

From Table~\ref{ablation_object_detection}, we empirically find that setting dilation to 3 ( the scale of anchors is 9 ) achieves the best detection performance when only one dilation is used, which may correspond to the dominate scale of objects in the dataset. We also observe a small performance gap among different dilations, which indicates the robustness of AADI. Moreover, since different dilations correspond to different scales of anchors, we expect that using two or more RPNs with different dilations will bring further performance gains. Therefore, we deploy two or more RPNs with different dilations in AADI and test the performance of these different dilation combinations. The results show that when two RPNs share the same parameters, there is almost no performance gain compared to using only one RPN. When the parameters are not shared, the performance can be substantially improved, and we can obtain the highest AP of 41.3 when the dilations of two RPNs are set to 2 and 4, respectively. On top of this, we add another RPN with dilation of 3 to test whether the performance can be further improved. However, a lower AP of 41.0 is obtained, probably due to newly added RPN parameters raising the difficulty of optimization. Moreover, compared with Faster R-CNN, our best practice of AADI achieves 3.4 AP gains. 

\begin{table}
  \begin{center}
  \resizebox{0.45\textwidth}{18mm}{
      \begin{tabular}{cccccccccccccccc}
          \toprule
          Dilation & RPN Parameter & AP & AP$_{50}$ & AP$_{75}$ & AP$_s$ & AP$_m$ & AP$_l$\\
          \midrule
          2  & - & 40.3 & 59.3 & 44.3 & 24.2 & 43.3 & 52.2\\
          3  & - & 40.8 & 59.5 & 45.0 & 24.0 & 44.6 & 53.1\\
          4  & - & 40.5 & 58.7 & 44.6 & 23.2 & 44.8 & 52.7\\
          \midrule
          2,3 & share & 40.8 & 59.6 & 45.2 & 24.4 & 44.0 & 53.0 \\ 
          3,4 & share & 40.7 & 59.2 & 44.5 & 23.3 & 44.6 & 54.0 \\
          2,4 & share & 40.6 & 59.0 & 44.5 & 23.7 & 44.0 & 52.4 \\
          \midrule
          2,3 & not share & 40.7 & 59.5 & 44.7 & \textbf{24.5} & 44.1 & 52.8\\
          3,4 & not share & 41.2 & 59.7 & \textbf{46.0} & 23.7 & 45.0 & 54.0\\
          2,4 & not share & \textbf{41.3} & \textbf{59.9} & 45.5 & 23.9 & \textbf{45.4} & \textbf{54.0}\\
          2,3,4 & not share & 41.0 & 59.6 & 45.1 & 24.4 & 44.7 & 53.6\\
          \midrule
          Faster R-CNN & - & 37.9 & 58.8 & 41.1 & 22.4 & 41.1 & 49.1\\
          \bottomrule
      \end{tabular}}
  \end{center}
  \caption{Detection performance of AADI for Faster R-CNN with different dilations. All results are trained with 1x schedule.}
  \label{ablation_object_detection}
\end{table}

\section{Conclusion}

In this paper, we propose a gradient-free anchor augmentation method named AADI. It does not add any parameters or hyper-parameters, and it does not significantly affect training and inference speed.
We conduct experiments on RetinaNet and different R-CNNs to verify its performance.
Specifically, AADI brings at least 2.2 box AP improvements on Faster/Mask R-CNN, 1.8 box AP improvements on RetinaNet, and 0.9 box AP gains on Cascade Mask R-CNN. In our future work, we will try to combine AADI and other anchor-based methods together to further improve detection performance.

\appendix

\section{The quality of anchors}

In our opinion, high quality anchors should have two characteristics. First, they have higher IoUs with ground truth to reduce the optimization difficulty of box regression.
Second, the positive and negative anchors can be easy to separate. We conduct several experiments to demonstrate these, specifically, for each ground truth box, we select an anchor with the highest IoU as its 'label', and the mean IoU of the 'label's of hand-designed anchors and augmented anchors are {\bf 0.5283} and {\bf 0.7662}, respectively. Furthermore, we regard the IoU of each anchor as its classification score, and the area under the receiver operating characteristic curve (AUC-ROC) of hand-designed anchors and augmented anchors are {\bf 0.8422} and {\bf 0.9417}, respectively, which shows that our augmented anchors (generated by AADI) are easier to separate.

\section{Dilated convolution in RPN}

For fair comparison, we also test 2x2 and 3x3 dilated convolution in RPN of Cascade R-CNN, where the baseline setting of dilation is 1. while the performance is almost same as baseline when dilation is 2, and slightly lower than baseline when dilation is 3. We hold the idea that the hand-designed anchors are hard to optimize, which will cause a large gradient, and the dilated convolution which enlarges the receptive field can not solve it. However, the RPN of AADI is trained on augmented anchors, and the augmentation process is gradient-free, which is beneficial for training.

\section{From the perspective of EM Algorithm}

For a standard EM algorithm, the variational evidence lower bound (ELBO) can be formulated as follows.

\begin{equation}
    \sum_i^n{logp(x_i; \theta)} \ge \sum_i^n\sum_z^Z{Q_i(z)log\frac{p(x_i, z; \theta)}{Q_i(z)}}
\end{equation}

For AADI, we can let anchors as $Z$, features extracted by backbone as $x$, and parameters of RPN as $\theta$. Moreover, RPN can be formulated as Eq.\ref{rpn_func}.
\begin{equation}
    b_i = f(x_i, \theta, z)
    \label{rpn_func}
\end{equation}
Where $b_i$ is proposal box generated by RPN.

Then, our likelihood function can be formulated as Eq.\ref{likelihood},

\begin{equation}
\label{likelihood}
\begin{aligned}
    p(x_i; \theta) &= IoU(b_i, y_i)\\
                   &= IoU(f(x_i, \theta, z), y_i)
\end{aligned}
\end{equation}
where $y_i$ is the corresponding target box. And $Q(Z)$ can be formulated as Eq.\ref{aadi_qz}.
\begin{equation}
    \label{aadi_qz}
    \begin{aligned}
        Q_i(z) &= p(z | x_i, \theta, \hat{z}) \\
               &= IoU(f(x_i, \theta, \hat{z}), y_i)
    \end{aligned}
\end{equation}
Where $\hat{z}$ can be hand-designed anchors or augmented anchors in previous iteration.

Therefore, in the E-step, we can fix the $\theta$ and estimate the hidden variable of $Z$ by RPN. And in the M-step, unlike standard EM algorithm, we can directly employ gradient descent algorithm to optimize the $\theta$ with estimated $Z$ and given $y_i$. As shown in Algorithm \ref{alg_aadi_formula}, line 2 can be viewed as the E-step, and line 3, 4 and 5 can be viewed as  M-step.

\section*{Acknowledgements}
This work was partially supported by The Key Program of National Natural Science Foundation of China under Grant (No. U1903213), the Natural Science Foundation of China (NSFC) (No. 61772296) and Shenzhen Fundamental Research Fund (No. JCYJ20170412170438636).

\bibliographystyle{named}
\bibliography{ijcai22}

\end{document}